\documentclass[sigconf]{acmart}

\usepackage[english]{babel}

\usepackage{graphicx}
\usepackage{booktabs} 
\usepackage{multirow}
\usepackage[normalem]{ulem}
\useunder{\uline}{\ul}{}

\setcopyright{rightsretained}

\acmDOI{10.475/978_1}

\acmISBN{978-1-4503-6615-1}

\acmConference[ICVGIP'21]{12th Indian Conference on Computer Vision, Graphics and Image Processing}{Dec. 2021}{Jodhpur, India}
\acmYear{2021}
\copyrightyear{2018}

\editor{Anoop M. Namboodiri}
\editor{Vineeth Balasubramanian}
\editor{Amit Roy-Chowdhury}
\editor{Guido Gerig}

\begin{document}

\title{DFW-PP: Dynamic Feature Weighting based Popularity Prediction for Social Media Content}


\author{Viswanatha Reddy G}
\affiliation{%
  \institution{IIIT Sri CIty}
}
\email{viswanatha.g15@iiits.in}

\author{Chaitanya B S N V}
\affiliation{%
  \institution{IIIT Sri CIty}
}
\email{viswachaitanya.b16@iiits.in}

\author{Prathyush P}
\affiliation{%
  \institution{IIIT Sri CIty}
}
\email{prathyush.p16@iiits.in}

\author{Sumanth M}
\affiliation{%
  \institution{IIIT Sri CIty}
}
\email{sumanth.m15@iiits.in}

\author{Mrinalini C}
\affiliation{%
  \institution{IIIT Sri CIty}
}
\email{mrinalini.c15@iiits.in}

\author{Dileep Kumar P}
\affiliation{%
  \institution{IIIT Sri CIty}
}
\email{dileepkumar.p15@iiits.in}

\author{Snehasis Mukherjee}
\affiliation{%
  \institution{Shiv Nadar University}
}
\email{snehasis.mukherjee@snu.edu.in}



\begin{abstract}
The increasing popularity of social media platforms makes it important to study user engagement, which is a crucial aspect of any marketing strategy or business model. The over-saturation of content on social media platforms has persuaded us to identify the important factors that affect content popularity. This comes from the fact that only an iota of the humongous content available online receives the attention of the target audience. Comprehensive research has been done in the area of popularity prediction using several Machine Learning techniques. However, we observe that there is still significant scope for improvement in analyzing the social importance of media content. We propose the DFW-PP framework, to learn the importance of different features that vary over time. Further, the proposed method controls the skewness of the distribution of the features by applying a log-log normalization. The proposed method is experimented with a benchmark dataset, to show promising results. The code will be made publicly available at \url{https://github.com/chaitnayabasava/DFW-PP}.
\end{abstract}

%


\maketitle
\pagestyle{plain}

\section{Introduction}

Popularity prediction of social media posts has been studied extensively in the past two decades, mostly due to its potential in commercial applications. Billions of pictures are downloaded and uploaded per second on the Internet through  social channels and photo-sharing mediums \cite{A.Zohourian}. However, not all of this content is noticed and admired by the desired viewers. Hence, finding out the important attributes (features) of a posted content that makes them popular may be a novel area of interest for research. For instance, applications like brand monitoring, political advertising or social media marketing are highly dependant on the characteristics of user interests. Notable efforts have been made to predict the popularity of social media content, understand its variations, and assess its growth \cite{szabo}, \cite{niu}, \cite{petrovic}, \cite{nwana}, \cite{pinto}, \cite{shamma}, \cite{reddy2020measuring}. However, due to the varieties in image content, limited efforts are found in the literature for assessing image content and predicting their popularity. In image popularity prediction, the main challenge lies with the number of views obtained by different individuals for a same photograph. As a result, applying models trained on the raw popularity scores would trigger a significant drop in the performance. This motivates us to develop a unified framework for image popularity dynamics.

Image Popularity can be defined as the level of interaction that a shared content receives on social websites (e.g., comments, shares, views and likes). The complexity of finding an index to measure popularity is quite high due to the diversity in the views received for the same image posted by different people. For example, views for the same image uploaded by celebrities would acquire likes in millions, whilst common people would receive only a few hundreds. This huge difference makes popularity prediction a challenging task. However, a rigorous analysis on the insightful patterns hidden within this massive data would aid in preserving users' interest on social platforms. There are various methodologies in discovering the correlated features inside the cluster of similar interests \cite{khosla}, \cite{ortis}, \cite{battiato}. Marketing agencies usually advertise their product to get the maximum reach by effectively utilizing the resources available in these social media platforms. However, popularity of the image is often variable with time. \citet{Cappallo} and \citet{gelli} estimated the popularity index by ignoring the popularity transformation over the time. In this study we emphasize on contextual as well as temporal information of the images posted in social media, to predict the popularity. Our contributions are three-fold:
\begin{itemize}
    \item A novel $log-log$ normalization technique is proposed to deal with the skew distribution of target variable.
    \item A novel framework is introduced based on dynamic weighting for different time periods of the input features. 
    \item Extensive experiments coupled with ablation studies are performed to show the impact of the proposed time based framework on popularity prediction.
\end{itemize}

Next we provide a survey of literature in the related topic.
\begin{figure}
\includegraphics[width=1.1\columnwidth]{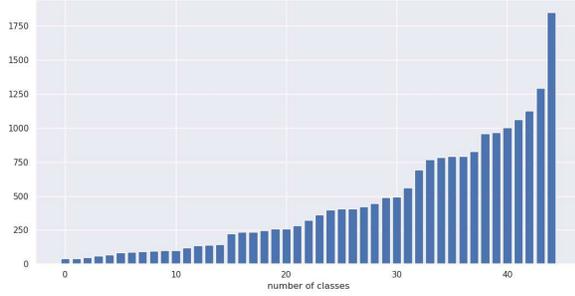}
  \caption{Distribution of samples per class for 45 clusters}
  \label{fig:longtail}
\end{figure}

 \begin{figure*}
\begin{center}
  \includegraphics[width=\linewidth]{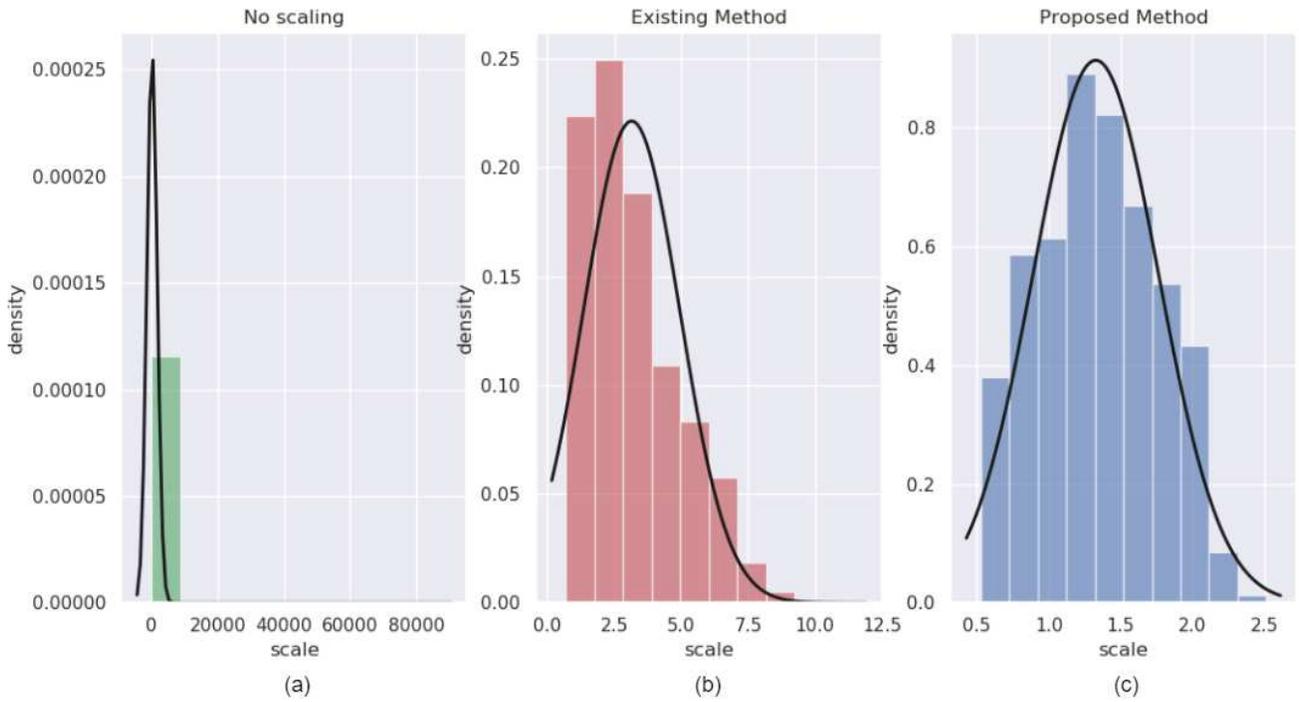}
\end{center}
\vspace{-.2in}
  \caption{Histographic representation of $s_{scale}$ using various scaling techniques. Existing method (b) corresponds to the method used in \cite{ortis}. Our proposed normalisation technique is shown in the (c), which follows a distribution closer to the normal.}
\label{fig:Fig1}
\end{figure*}

\section{Related Works}
Popularity prediction in social media contents has been an emerging research area during the last few years. Current research trends on predicting popular social posts are based on text, image, or a combination of both. Few studies even considered the time factor.
Image Popularity dynamics prediction techniques differ in the means of various metrics. Metrics can be any of the following : View(s), share(s) or Re-share(s), Comment(s) and Count(s). Moreover, these metrics are collected from the basic pipeline used for the extraction and evaluation of various features impacting the popularity metrics, followed by a Machine Learning Model to predict the popularity. As a result, we analyse these works by categorizing them based on the features and the Machine Learning algorithms.

The conventional methods for image popularity prediction were mostly focused on the low level features such as image tags. \citet{yamasaki} estimated the popularity of Flickr\footnote{\url{https://www.flickr.com/}} images using tags. The importance of each tag is predicted based on the tag frequency and the weights. The features obtained are used for, either prediction of number of views, likes, comments (regression) or differentiation between popular and unpopular images (classification).

Zohourian et al. \cite{A.Zohourian} examined the popularity of online posts on the Instagram with three unique business profiles. They normalised the number of views with the number of followers for a given profile. With regard to the business profile analysis, the count of followers can be increased in a very short span of time. Aloufi et al. \cite{aloufi} predicted the levels of some predefined interaction by combining distinct features and inspecting the impact of those combinations. Interactions can be as follows: number of comments, number of views and number of favourites on Flickr.

Fewer attempts have been made to analyze the image content for popularity prediction. \citet{khosla} proposed a popularity score obtained by taking the cumulative engagement of users up to the download time and normalizing it with the number of days after which the content was posted. They investigated basic image characteristics such as intensity variance, color and low-level vision features including gist, texture, color patches and gradient for images. Moreover, high-level image features like existence of different objects were considered to predict the number of views using Linear Support Vector Regressor.
\begin{figure*}
\begin{center}
  \includegraphics[width=\linewidth]{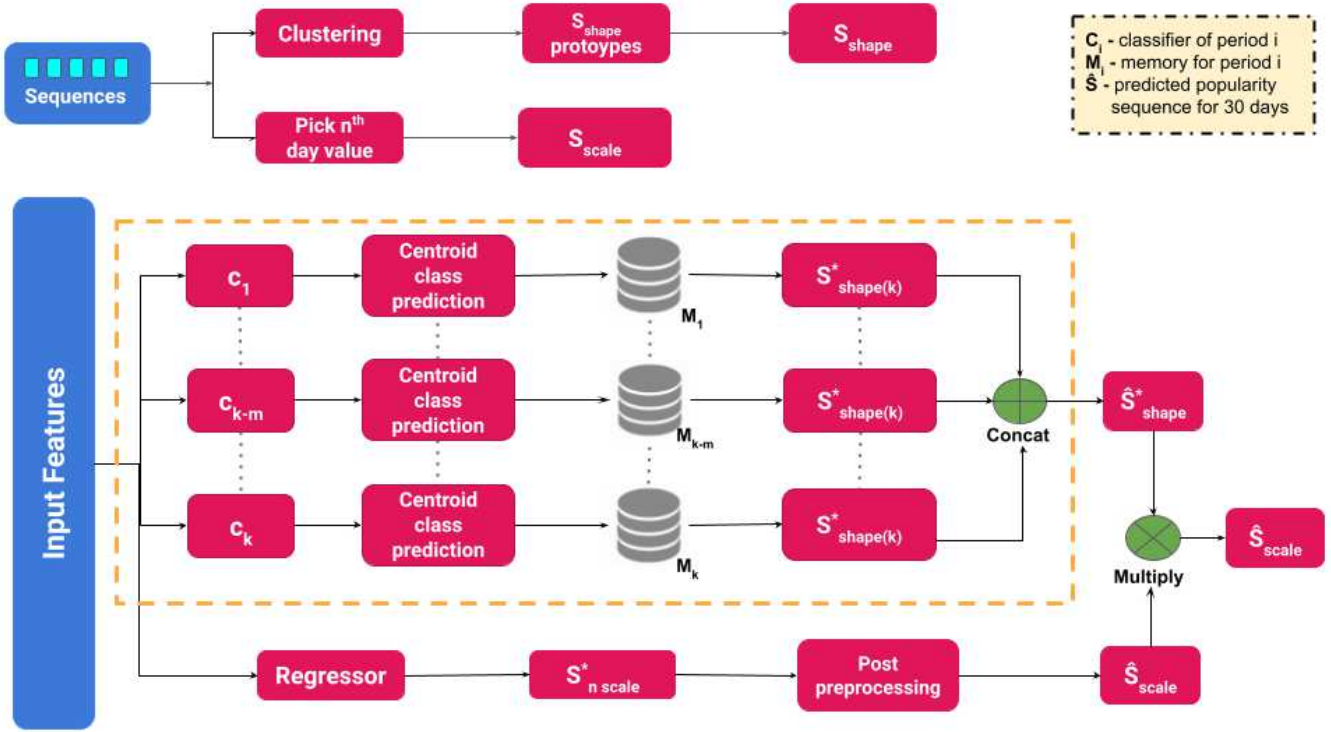}
\end{center}
\caption{Proposed framework for the popularity sequence prediction task}
\label{fig:Fig3}
\end{figure*}

\citet{gelli} applied visual sentiment features, object features, context features and user features to predict the popularity metric of social images using SVM and CNN. \citet{McParlane} combined the picture contents, frame contexts, user context and user tags to predict the number of views and comments. Cappallo et al. \cite{Cappallo} demonstrated the problem of popularity prediction using Latent SVM. They considered the task as a ranking problem. The cost function was trained to conserve the ranking of the popularity scores of the images. Number of views and comments were used to find the score for Flickr Images, whereas for Twitter, they considered the count of re-tweets and favourites.


Hu et al. \cite{Hu} implemented a Caffe \cite{caffe} deep learning framework on Yahoo Flickr Creative Commons 100M (YFCC100M)\footnote{\url{https://yahooresearch.tumblr.com/post/89783581601/one-hundred-million-creative-commons-flickr-images-for}} to extract the visual features for each image and compared several multimodal and unary learning approaches to predict the popularity. They used tag features along with visual features of the images. They concluded that the tag features outperform all other unimodal and multimodal learning models. Presumably very few works focused on taking the popularity dynamics into consideration. Wu et al. \cite{Wu} presented a multi-time scale representation of popularity by modelling time sensitive contexts. They demonstrated that the time of post plays an important role in popularity prediction.

Motivated by \citet{Wu}, \citet{ortis} utilized the time information of images in predicting popularity. Following \citep{ortis}, we focus on the time of posting of the images for predicting the popularity. They assign the same weight on images posted on Day 1 to Day 30. In reality, the popularity of an image is expected to go down as days progress. With this assumption, we propose a dynamic weighting scheme which takes into account the number of days from posting.

\section{PROPOSED METHOD}
The proposed method for image popularity prediction is based upon the image content and the relevance of the image according to time.

\subsection{Motivation}
The motivation behind the proposed method is \citet{ortis}, which shows state-of-the-art performance in terms of being able to predict the popularity sequence of an image right from Day 0. However, we identified the following three limitations of \citet{ortis}, and proposed necessary changes accordingly, in order to enhance the performance. 

\subsubsection{Constant Weightage:} \citet{ortis} used the same weight for each feature across the time-frame, making the model insensitive to the change of feature(s) importance across time. We believe that each feature contributes differently in different time frames. For instance, contextual information of an image may suggest higher popularity score during the initial time frame (say Day 1) but, it may not contribute similarly as days progress.

\subsubsection{Clustering:} \citet{ortis} clustered the temporal (daily) sequence of popularity scores of the images  into 45 classes. However, we observe (Figure \ref{fig:longtail}) that the 45 classes suffer from a huge class imbalance. Hence, we experimentally reduced the number of clusters so that the class imbalance problems are mitigated.

\subsubsection{Skew Distribution:} Finally, the skewness in the distribution of a variable for measuring the importance score, is another limitation in \cite{ortis}, as shown in Figure \ref{fig:Fig1}, here $s_{scale}$ is the sequence scale which we further elaborate in section \ref{problemdef}. We propose a log-log normalization on the importance scores, to reduce the problem of skewness.

We start by establishing the problem definition, followed by the proposed methodologies.

\subsection{Problem Definition}
\label{problemdef}
Given the number of views achieved by a Flickr photo over $n$ days, we define the engagement sequence $s$ as
\begin{equation}
\label{sequence}
s = [d_{0}, d_{1}, ..., d_{n}],
\end{equation}
where $s \in \mathbb{R}^{n}$ and $d_{i}$ is the popularity score achieved as of day $i$. Following \citet{ortis}, $s$ is further split into two parts:

\textbf{sequence scale ($s_{scale}$)} is defined as the maximum element of the vector $s$, i.e., number of views till the $n^{th}$ day. This means,
\begin{equation}
\label{scale}
s_{scale} = max(s) = d_{n}.
\end{equation}

\textbf{sequence shape ($s_{shape}$)} is obtained by dividing the vector $s$ by the scale ($s_{scale}$).
\begin{equation}
\label{shape}
s_{shape} = \dfrac{s}{s_{scale}} 
\implies
s_{shape} = \left[\dfrac{d_{0}}{d_{n}}, \dfrac{d_{1}}{d_{n}}, ..., \dfrac{d_{n}}{d_{n}}\right].
\end{equation}

The vector $s$ represents a cumulative function, so that $d_{n}$ always corresponds to the maximum value of $s$. Therefore from (\ref{shape}), we can consider $s$ as the combination of scale ($s_{scale}$) and shape ($s_{shape}$) attributes as given by:
\begin{equation}
\label{maineq}
s = s_{shape} * s_{scale}.
\end{equation}

The scale represents the popularity level of the image w.r.t the total engagement after $n$ days. The shape generalizes the temporal history (i.e., trend) of the popularity of image across the observation time, independent of its actual values. Similar to \citet{ortis}, we assume an independent association between the two attributes. For instance, two sequences having the same shape might have quite distinct scales. This assumption allows us to decouple the scale and shape learning to predict them independently before finally estimating $s$.

\subsection{Proposed Shape Prototyping}
According to the formulation given in (\ref{shape}), all the
values of $s_{shape}$ sequences are in the range $[0, 1]$. We assume sequences with the same dynamics to have a relatively similar $s_{shape}$. First, we identify a number of engagement prototypes representing various shape groups since groups of sequences with same shape are examples of dynamics with common engagement evolution. Specifically, we use the K-means \cite{hartigan1979algorithm} algorithm to cluster the normalized sequences (i.e., $s_{shape}$). In order to discover the optimal $K$ value for K-means algorithm, we make use of the elbow method. 

After clustering, we get a collection of shape prototypes interpreted as a memory ($M$) of temporal dynamics for popularity sequences. Here, memory $M \in \mathbb{R}^{Kxn}$, where $K$ corresponds to number of prototypes (clusters), and $n$ represents prototype dimension (length) of the period. All sequences are allocated to a cluster and are represented by their respective shape prototype ($s_{shape}^{*}$). This problem can be treated as a $K-way$ classification. The training set is used to build a classifier that predicts the prototype class (shape of the matching sequence) based only on the features collected from a social post. The prototype class will be used to retrieve the shape prototype $s_{shape}^{*}$ from the memory $M$.

Figure \ref{fig:Fig3} shows the overall proposed framework for the popularity sequence prediction task.



\subsection{Updated Scale Estimation}
The scale is estimated using a regressor model. \citet{ortis} used the normalizing scheme shown in (\ref{originalnorm}), which resulted in a highly skewed distribution, as shown in Figure \ref{fig:Fig1}b. In skewed data, the tail region may act as an outlier and adversely affect the performance, especially for regression-based models. To alleviate this issue, we propose a log-log normalization scheme, which transforms the skew data to have a normal distribution, as shown in Figure \ref{fig:Fig1}c. 
\begin{equation}
\label{originalnorm}
s_{nscale} = log\left(\dfrac{s_{scale}}{n} + 1\right),
\end{equation}
\begin{equation}
\label{proposedscheme}
s_{nscale} = log(log(s_{scale} + c) + c),
\end{equation}
where $s_{nscale}$ is the normalized scale value and $n$ is equal to 30 (number of days) and $c$ is any positive number, to avoid zero value for the log function. We set $c=1$ for all our experiments. Hence,
\begin{equation}
\label{denormalized:log-log}
\hat{s}_{scale} = e^{(e^{\hat{s}_{nscale}} - c)} - c.
\end{equation}
\begin{table}
\caption{Number of features engineered per category.}
\label{tab:feature_importance}
\begin{tabular}{|c|c|}
\hline
\textbf{Feature Category} & \textbf{\begin{tabular}[c]{@{}c@{}}Number of features \\ engineered\end{tabular}} \\ \hline
Text & 15 \\ \hline
User & 13 \\ \hline
Image & 5 \\ \hline
Time & 4 \\ \hline
\end{tabular}
\end{table}

The trained regression model estimates the normalized scale ($s_{nscale}$). We use (\ref{denormalized:log-log}) to transform this predicted normalized scale ($\hat{s}_{nscale}$) to get the total no of views on the $n^{th}$ day. 
\begin{figure*}
\includegraphics[width=0.9\textwidth]{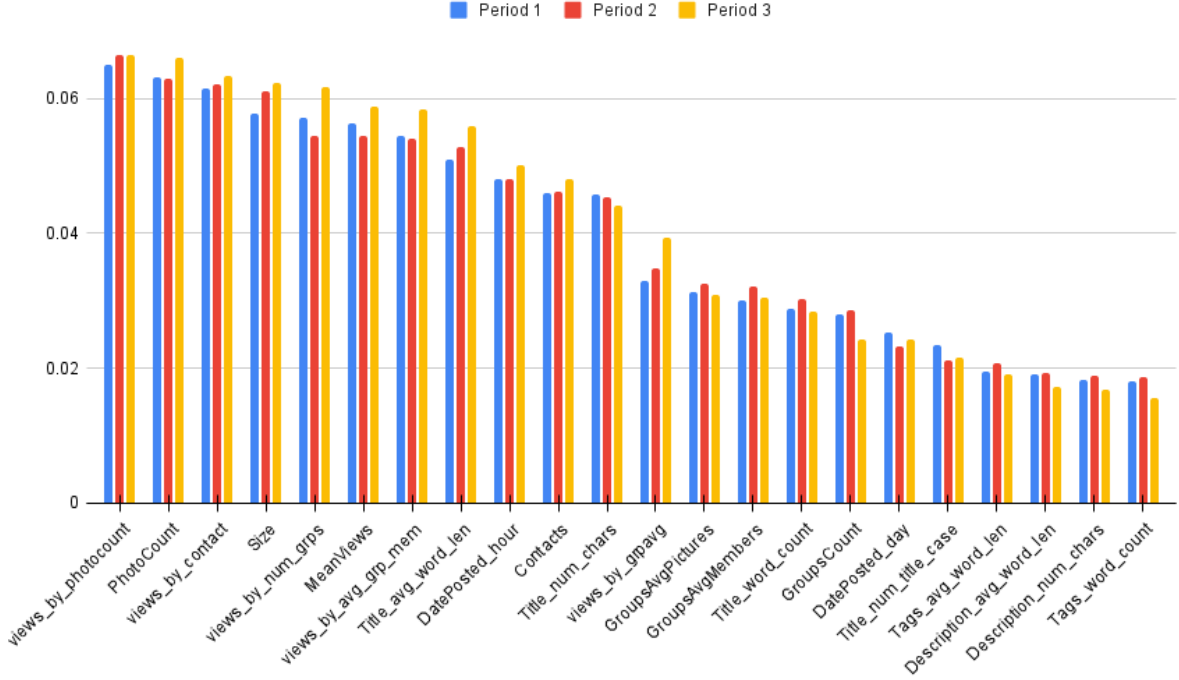}
  \caption{$s_{shape}$ Random Forest Classifier Feature Importance's under different periods.}
  \label{fig:feature_importance_pl}
\end{figure*}

\subsection{Dynamic Feature Weighting}
We propose a dynamic feature weighting framework to assign dynamic weights to each considered input feature, w.r.t the sub-period. Given an input sequence of time period $p$ of length $n$, we divide the total time period into $k$ equal length sub-periods ($p=[p_{1}, p_{2}, ..., p_{k}]$), each with a length of $\dfrac{n}{k}$. The popularity sequence ($s_{i}$) corresponding to the sub-period $p_{i}$ is given by,
\begin{equation}
\label{subperiodseq}
s_{i} = s\left[(i-1)*\dfrac{n}{k}: i*\dfrac{n}{k}\right].
\end{equation}

While predicting the shape prototype $\hat{s}_{shape_i}^{*} \in \mathbb{R}^{n/k}$ for the entire time frame, each sub-period $p_{i}$ has access to an isolated classifier model ($C_{i}$) and memory $M_{i} \in \mathbb{R}^{K_{i}x\dfrac{n}{k}}$, where $K_{i}$ is the number of prototypes for the sub-period $p_{i}$. In the proposed framework, for estimating the shape prototypes, we effectively provide dynamic weights for features in different periods, so the priority assigned to the input features will not be the same in each period. We use a single regressor model for predicting $\hat{s}_{scale}$, the scale attribute of sequence ($s$), which is defined as the popularity score after $n^{th}$ ($n = 30$) day and so doesn't alter from period to period.

\begin{table*}
\centering
\caption{The performance of the proposed framework in terms of the Case C \& D metrics as compared to the existing $s_{scale}$ normalization techniques.}
\label{table:normalizationresultsCD}
\begin{tabular}{|c|c|c|c|c|c|c|c|c|} 
\hline
\multirow{2}{*}{\begin{tabular}[c]{@{}c@{}}\textbf{Scaling}\\\textbf{Method}\end{tabular}} & \multicolumn{4}{c|}{\textbf{Case C}} & \multicolumn{4}{c|}{\textbf{Case D}} \\ 
\cline{2-9}
 & $tRMSE_{mean}$ & $tRMSE_{median}$ & $tMAE_{mean}$ & $tMAE_{median}$ & $tRMSE_{mean}$ & $tRMSE_{median}$ & $tMAE_{mean}$ & $tMAE_{median}$ \\ 
\hline
No scaling & 14.387±0.131 & 9.944±0.092 & 14.196±0.138 & 9.798±0.094 & 14.331±0.13 & 9.888±0.101 & 14.135±0.138 & 9.752±0.085 \\ 
\hline
log & 9.108±0.083 & 6.487±0.021 & 8.892±0.084 & 6.339±0.034 & 9.467±0.084 & 6.83±0.016 & 9.245±0.079 & 6.683±0.018 \\ 
\hline
\begin{tabular}[c]{@{}c@{}}log(scale/30 \\+ 0.1)\end{tabular} & 9.327±0.075 & 6.77±0.066 & 9.11±0.076 & 6.6±0.061 & 9.724±0.069 & 7.136±0.043 & 9.503±0.067 & 6.973±0.028 \\ 
\hline
\citet{ortis}&  10.535±0.088 & 8.043±0.036 & 10.302±0.086 & 7.856±0.04 & 11.073±0.113 & 8.532±0.028 & 10.836±0.112 & 8.377±0.043 \\ 
\hline
\begin{tabular}[c]{@{}c@{}}Proposed \\normalization\end{tabular} & \textbf{8.773±0.053} & \textbf{6.066±0.009} & \textbf{8.564±0.056} & \textbf{5.927±0.015} & \textbf{9.048±0.077} & \textbf{6.362±0.036} & \textbf{8.831±0.082} & \textbf{6.205±0.041} \\
\hline
\end{tabular}
\end{table*}

\begin{table*}
\centering
\caption{The performance of the proposed $s_{scale}$ normalization technique in terms of the Case C \& D metrics for different values of c. Our normalization scheme is invariant to c value compared to the normalization scheme used in \cite{ortis}}
\label{table:normalizationresultsforc}
\begin{tabular}{|c|c|c|c|c|c|c|c|c|} 
\hline
\multirow{2}{*}{\begin{tabular}[c]{@{}c@{}}\textbf{Scaling}\\\textbf{Method}\end{tabular}} & \multicolumn{4}{c|}{\textbf{Case C}}                                                                  & \multicolumn{4}{c|}{\textbf{Case D}}                                                                                                                                           \\ 
\cline{2-9}
                                                                                            & $tRMSE_{mean}$         & $tRMSE_{median}$     & $tMAE_{mean}$         & $tMAE_{median}$     & $tRMSE_{mean}$         & $tRMSE_{median}$     & $tMAE_{mean}$        & $tMAE_{median}$                                                                           \\ 
\hline


c = 0.5                                                                                      & ~8.782±0.063~           & ~6.032±0.043~           & ~8.574±0.074~          & ~5.897±0.063~           & ~9.035±0.086~             & ~6.269±0.015~           & ~8.817±0.094~          & ~6.119±0.006~                                                                                   \\ 
\hline

c = 0.1                                                                                    & ~8.961±0.057~           & ~6.112±0.063~           & ~8.749±0.063~          & ~5.958±0.063 ~           & ~9.147±0.057~             & ~6.28±0.049~           & ~8.926±0.062~          & ~6.122±0.041~                                                                                   \\ 
\hline


\begin{tabular}[c]{@{}c@{}} c =1  \end{tabular}                                 & \textbf{8.773±0.053} & \textbf{6.066±0.009} & \textbf{8.564±0.056} & \textbf{5.927±0.015} & \textbf{~9.048±0.077} & \textbf{6.362±0.036} & \textbf{8.831±0.082} & \textbf{6.205±0.041}   \\
\hline

\begin{tabular}[c]{@{}c@{}}log(scale/30 \\+ 0.1)\end{tabular} & 9.327±0.075 & 6.77±0.066 & 9.11±0.076 & 6.6±0.061 & 9.724±0.069 & 7.136±0.043 & 9.503±0.067 & 6.973±0.028 \\ 
\hline
\begin{tabular}[c]{@{}c@{}}log(scale/30\\+ 0.5)\end{tabular} & 10.036±0.136 & 7.498±0.018 & 9.812±0.128 & 7.331±0.016 & 10.523±0.129 & 7.974±0.046 & 10.296±0.127 & 7.801±0.03\\ 
\hline
 \begin{tabular}[c]{@{}c@{}}log(scale/30\\+ 1) \cite{ortis}\end{tabular} & 10.535±0.088 & 8.043±0.036 & 10.302±0.086 & 7.856±0.04 & 11.073±0.113 & 8.532±0.028 & 10.836±0.112 & 8.377±0.043 \\ 
\hline

\end{tabular}
\end{table*}

\subsection{Inference}
The input features are fed to the regressor to get the normalized scale estimate ($\hat{s}_{nscale}$), which is post-processed using (\ref{denormalized:log-log}) to get $\hat{s}_{scale}$. Then the classifiers of the $k$ sub-periods are fed with the input features to get the respective shape prototype classes. For each sub-period, using these predicted classes and their respective memory ($M_{i}$), we obtain the ($\hat{s}^{*}_{shape_i}$) each of length $\dfrac{n}{k}$. We then concatenate these sub-period shape prototypes to get the overall shape attribute for $n$-days as follows:
\begin{equation}
\label{shapeconcat}
\hat{s}_{shape}^{*} = Concat(\hat{s}_{shape_1}^{*}, \hat{s}_{shape_2}^{*}, ..., \hat{s}_{shape_k}^{*}),
\end{equation}
where $\hat{s}_{shape_i}^{*} \in \mathbb{R}^{n/k}$ is the predicted shape prototype of period $p_{i}$ and $\hat{s}_{shape}^{*} \in \mathbb{R}^{n}$ is the shape prototype for the overall period. We then predict the final popularity sequence as follows:
\begin{equation}
\label{finalseq}
\hat{s} = \hat{s}_{scale} * \hat{s}_{shape}^{*}
\end{equation}


\subsection{Framework Evaluation:}
To examine the influence of these independent components (shape and scale) on dynamic popularity estimation, we consider the three experimental settings used in \cite{ortis}. Furthermore, we propose an additional case to get a generalized outlook of the performance of the proposed framework. The evaluation settings are defined as follows:

$\boldsymbol{CASE-A}$: In this setting, the combination of ground truth shape prototype ($s_{shape}^{*}$) assigned by the clustering procedure and the ground truth scale value ($s_{scale}$) are used to obtain the output sequence. This case achieves the minimum possible error (upper bound) and the errors are due to the clustering approximation of the sequences.

$\boldsymbol{CASE-B}$: In this setting, the ground truth scale values $s_{scale}$ are considered, whereas the shape prototype $\hat{s}_{shape}^{*}$ is estimated using a classifier. The final error in this case is due to the sequence cluster approximation and the error of the classifier used to predict the shape prototype.

$\boldsymbol{CASE-C}$: Here we exploit the ground truth shape prototype $s_{shape}^{*}$, but use the regressor and (\ref{denormalized:log-log}) to predict the scale value $\hat{s}_{scale}$. The error, in this case, can be ascribed to clustering approximation and scale attribute estimation error by the regression model.

$\boldsymbol{CASE-D}$: In this case, we combine the scale value $\hat{s}_{scale}$ obtained using the regressor and the shape prototype $\hat{s}_{shape}^{*}$ obtained using the classifier. This case helps to measure the overall performance of the framework when the predictions entirely come from the models.


Following \citet{ortis}, we compare the model performance in all the $4$ cases using truncated RMSE mean ($tRMSE_{mean}$) and median ($tRMSE_{median}$). Additionally, we report truncated MAE mean ($tMAE_{mean}$) and median ($tMAE_{median}$). The $tRMSE$ and $tMAE$ are robust to outliers, as we would ignore the best and the worst $25\%$ values from RMSE and MAE, resulting in a much clearer distribution for the model evaluation.
\begin{table*}[]
\caption{The performance in terms of the Case C \& D metrics for different time period lengths}
\label{table:caseCDtime}
\begin{tabular}{|c|c|c|c|c|c|c|c|c|}
\hline
\multirow{2}{*}{\textbf{\begin{tabular}[c]{@{}c@{}}Period \\ Length\end{tabular}}} & \multicolumn{4}{c|}{\textbf{Case C}} & \multicolumn{4}{c|}{\textbf{Case D}} \\ \cline{2-9} 
 & $tRMSE_{mean}$  & $tRMSE_{median}$  & $tMAE_{mean}$  & $tMAE_{median}$  & $tRMSE_{mean}$  & $tRMSE_{median}$  & $tMAE_{mean}$  & $tMAE_{median}$  \\ \hline
5 & 8.572±0.056 & 5.942±0.068 & 8.524±0.087 & 5.887±0.073 & 8.906±0.072 & 6.271±0.035 & 8.847±0.069 & 6.208±0.042 \\ \hline
10 & 8.782±0.056 & 6.105±0.055 & 8.573±0.062 & 5.976±0.063 & 9.06±0.056 & 6.372±0.031 & 8.843±0.061 & 6.223±0.039 \\ \hline
15 & 8.943±0.04 & 6.179±0.05 & 8.644±0.051 & 5.969±0.042 & 9.228±0.078 & 6.438±0.069 & 8.909±0.087 & 6.215±0.055 \\ \hline
30 & 9.248±0.225 & 6.398±0.12 & 8.738±0.203 & 6.068±0.1 & 9.578±0.21 & 6.656±0.49 & 9.009±0.19 & 6.257±0.038 \\ \hline
\end{tabular}
\end{table*}

\begin{table*}
\centering
\caption{The performance in terms of the Case C \& D metrics for different combinations of cluster sizes}
\label{table:mainresults}
\begin{tabular}{|c|c|c|c|c|c|c|c|c|} 
\hline
\multirow{2}{*}{ \textbf{Cluster Size} } & \multicolumn{4}{c|}{\textbf{Case C} } & \multicolumn{4}{c|}{\textbf{Case D} } \\ 
\cline{2-9}
 & $tRMSE_{mean}$  & $tRMSE_{median}$  & $tMAE_{mean}$  & $tMAE_{median}$  & $tRMSE_{mean}$  & $tRMSE_{median}$  & $tMAE_{mean}$  & $tMAE_{median}$  \\ 
\hline
(2, 2, 2) & 8.773±0.053   &   6.066±0.009 & 8.564±0.056 & 5.927±0.015 & 9.048±0.077   &   6.362±0.036 & 8.831±0.082 & 6.205±0.041 \\
\hline
(2, 2, 3) &	8.756±0.058 & 6.046±0.018 & 8.549±0.062 & 5.916±0.027 &	9.047±0.058 & 6.332±0.03 & 8.827±0.062 & 6.184±0.026 \\
\hline
(3, 3, 3) &	8.699±0.072 & 6.049±0.029 & 8.489±0.072 & 5.901±0.039 &	9.025±0.069 & 6.332±0.053 & 8.801±0.066 & 6.186±0.049 \\
\hline
(3, 3, 4) &	8.693±0.041 & 6.016±0.019 & 8.49±0.038 & 5.871±0.009 &	9.001±0.037 & 6.312±0.06 & 8.777±0.032 & 6.153±0.057 \\
\hline

(3, 2, 4) &	8.739±0.031 & 6.037±0.035 & 8.53±0.027 & 5.892±0.036 &	9.051±0.035 & 6.341±0.04 & 8.827±0.026 & 6.19±0.022 \\
\hline
(3, 2, 3) &	8.699±0.061 & 6.063±0.016 & 8.493±0.058 & 5.911±0.018 &	9.02±0.041 & 6.377±0.025 & 8.803±0.037 & 6.231±0.021 \\


\hline
(3, 3, 5) &  8.686±0.103 & 6.03±0.069 & 8.484±0.106 & 5.893±0.081 &	9.01±0.108 & 6.338±0.049 & 8.784±0.112 & 6.186±0.05	\\																			
\hline
\end{tabular}
\end{table*}
\begin{table*}
\centering
\caption{The performance in terms of the Case C \& D metrics for different classification and regression modeling techniques. Note that the result of existing method is taken from the respective paper.}
\label{table:mainresultsCD}
\begin{tabular}{|c|c|c|c|c|c|c|c|c|} 
\hline
\multirow{2}{*}{ \textbf{Model} } & \multicolumn{4}{c|}{\textbf{Case C} } & \multicolumn{4}{c|}{\textbf{Case D} } \\ 
\cline{2-9}
 & $tRMSE_{mean}$  & $tRMSE_{med}$  & $tMAE_{mean}$  & $tMAE_{med}$  & $tRMSE_{mean}$  & $tRMSE_{med}$  & $tMAE_{mean}$  & $tMAE_{med}$  \\ 
\hline
XGBoost  & 9.736 ± 0.098 & 6.721±0.054 & 9.522±0.099 & 6.57±0.05 & 10.08±0.089 & 7.045±0.025 & 9.856±0.096 & 6.891±0.03 \\ 
\hline
LightGBM & 9.391 ± 0.087 & 6.515±0.067 & 9.181±0.091 & 6.378±0.06 & 9.737±0.084 & 6.864±0.068 & 9.518±0.083 & 6.706±0.06 \\ 
\hline
Bayesian & 11.626 ± 0.144 & 8.008±0.058 & 11.458±0.15 & 7.899±0.065 & 11.95±0.16 & 8.356±0.053 & 11.778±0.166 & 8.241±0.048 \\ 
\hline
SVM  & 10.53±0.048~ & ~7.218±0.009~ & 10.331±0.051 & 7.086±0.011~ & ~10.914±0.056 & 7.635±0.033 & 10.711±0.06 & 7.511±0.036~ \\ 
\hline
Resnet 1D & 10.288 ± 0.266 & 6.977 ± 0.3 & 10.131 ± 0.273 & 6.884 ± 0.283 & 10.531 ± 0.349 & 7.421 ± 0.404 & 10.372 ± 0.356 & 7.32 ± 0.393 \\ 
\hline
\citet{ortis} & 9.37 & 7.38 & - & - & - & - & - & - \\ 
\hline
Random Forest & \textbf{8.773±0.053}  & \textbf{6.066±0.009}  & \textbf{8.564±0.056}  & \textbf{5.927±0.015} & \textbf{9.048±0.077} & \textbf{6.362±0.036}  & \textbf{8.831±0.082}  & \textbf{6.205±0.041}  \\
\hline
\end{tabular}
\end{table*}

\section{Experiments and Ablation Studies}
\label{ablation}

\textbf{Experimental Setups: } We validate the efficacy of the proposed framework using a set of comprehensive experiments and study their characteristics through each of the four evaluation settings. We avoid substantial hyper-parameter tuning due to a large number of chosen experimental combination settings. The reported metrics can serve as a baseline for further extensions. We are able to achieve better metrics compared to \cite{ortis} even without much hyper-parameter tuning showing the effectiveness of our proposed approach. All the reported metrics can further be improved by extensive hyper-parameter tuning.


We conduct our experiments using 3-fold validation. The mean and standard deviation are reported for statistical significance. For all the experiments, we use K-means for two way clustering and random forest model for both regression and classification. The number of periods is set to three, and we make use of feature engineering. We evaluate the model using the Cases A, B, C \& D. Due to space constraints we report Case C \& D results in the main paper, however please refer to the Appendix for Case A \& B results. 

\textbf{Dataset:} We conduct experiments on the popularity dataset from \cite{ortis} which was collected from the Flickr API. The dataset consists of 21,035 images and is monitored at regular intervals of 10, 20 and 30 days.

\subsection{Feature Engineering and Analysis}
To improve the popularity prediction score, we explore a variety of feature combinations. The features we experiment with, are from the following four categories: textual, user, image and time features. For the user features category, we engineer a new feature set by dividing MeanViews (average views) by PhotoCount, NumGroups (number of groups), Contacts, GroupsAvgPictures (average number of photos uploaded by the group) and AvgGroupsMemb. In order to create a textual feature set, pre-processing techniques have been employed on text data. These include WordCount, Length, AverageWordLength, CountOfUpperCaseWords, and CountOfTitleCaseWords. Next, for the time features set we engineer features using seasons (spring, winter, summer, fall), times of day (breakfast, lunch, dinner preparation, dinner, desserts), year, month, hour and day of the week. The counts of all considered and developed features in each category are listed in Table \ref{tab:feature_importance}. All these engineered features along with the original features are used as inputs, which effectively make the input size as $37$ for all the considered experiments.


A feature importance study on a 30-day time frame is carried out to determine the usefulness of the engineered features. We can observe that most of the top features are from the set of newly engineered features. views\_by\_photocount, DatePosted\_hour and Title\_avg\_word\_len are few of the top features which shows the importance of variety and contribution of different categories in the engineered features. Also, it is clear from the figure \ref{fig:feature_importance_pl} that user-related features are among the most important features for all time periods. We observe that the proposed Dynamic Feature Weighting is supported by the variations in feature significance across time periods. Further, we infer that for the top features, importance contribution is higher for the third period as compared to the first two periods. This reduces gradually for the lower-ranked features for which the feature importance contribution is higher in the first two periods than in the third.

\subsection{Proposed Normalization Scheme}

We validate the effectiveness of the proposed scale normalization scheme shown in (\ref{proposedscheme}). For a fair comparison, we trained and tested all the different normalization schemes under the same settings(using best model combination). Table \ref{table:normalizationresultsCD} shows that the proposed scale normalization scheme (\ref{proposedscheme}) has significantly boosted the framework's performance. When compared to the case where no scaling scheme is applied, we can observe about 37\% reduction in $tRMSE_{mean}$ for Case-D through our scheme and about 23\% reduction in the same when compared with \cite{ortis}. This shows that our proposed $log-log$ transformation effectively mitigate the skew distribution issues. Our strategy has outperformed the existing scaling methods in all the metrics.

Further, we show in Table \ref{table:normalizationresultsforc} that results achieved by the suggested method are invariant with respect to the constant value ($c$) used in (\ref{proposedscheme}), but the normalization strategy utilized in \cite{ortis} is considerably impacted. 
The Case D $tRMSE_{mean}$ is 11.07 when c is set to 1, and it lowers to 9.724 when c=0.1, but metrics for our technique are constant throughout a wide range of c's value. Meaning, the advantage of our transformation becomes more clear compared to existing work, indicating a superior transformation to cope with skewness. We can also conclude that our scheme offers an edge to reduce the extensive parameter tuning for the constant.



\subsection{Time period Analysis}
\label{timeperiod:exp}
We experiment with different period lengths ($p_{len}$) of 5, 10, 15, 30 and report the metrics in Table \ref{table:caseCDtime} for cases C \& D. The results shows that partitioning the sequence into smaller period lengths can affect the final predictions performance. We observed a diminution in performance as the period length increased from 5 to 30, emphasizing the effect of assigning dynamic weights to the features in each period ($p_{i}$).

\begin{equation}
\label{periodlength}
p_{len} = \dfrac{n}{k} \implies k = n * p_{len},
\end{equation}
where $p_{len}$ is the length of each sub-period. $n$ is the overall period's sequence length (which is 30 in all the experiments), and $k$ is the number of sub-periods each of length $p_{len}$ .


For each period length $p_{len}$ considered, we divide the entire period of n days into sub-periods of equal length $p_{len}$. For each sub-period, we trained a classifier, which results in a total of k classifiers, one for each of the considered period lengths.

From Table \ref{table:caseCDtime}, we infer that using the period length of 5 helped to achieve the best performance. However, it would require six classifiers, one for each period and one regressor in total. So, we chose to utilize a period length of 10, which gave an almost similar performance in all 4 cases. This setting would require only three classifiers and one regressor, effectively reducing the number of models to train to half while resulting in a minimum loss in performance.


\subsection{Analysis on number of clusters}
\label{clsuter}
To determine the optimal number of clusters in the K-means algorithm, we apply the elbow method. Following the cluster setting of \citet{ortis}, we implemented the elbow method experiment for $1$ to $50$ cluster sizes. From these elbow method graphs (refer to the Appendix Fig~\ref{fig:elbow_50} \& Fig~\ref{fig:elbow_10}), we can infer that the optimal $K$ value is lying in the range of [1, 10]. Hence, we further implemented the elbow method experiments for $1$ to $10$ cluster sizes to get a finer optimal range for the $K$ value. We then carryout additional experiments with different combinations of cluster sizes in this optimal range for the $3$ periods.

From Table \ref{table:mainresults}, we can infer that though the optimal $K$ value for period-1, period-2 and period-3 are 3, 3, and 4 respectively, but the difference in results across various combinations of the cluster sizes is almost negligible. So, we utilised the $(2, 2, 2)$ cluster settings throughout the study for simplicity.




\subsection{Models for Regression and Classification}
We experimented with various Machine learning and Deep learning models for the classification and regression tasks.

    \textbf{Tree Based Models}: The Tree-based models like regression trees, random forest and gradient boosted trees are widely used for various machine learning tasks. They help in achieving high prediction accuracy and model selectivity. The random effect helps to properly utilize the related correlation structure and improves the performance of the model by allowing statistical inference \cite{friedman2001elements}. We experiment with Random Forest, LightGBM \cite{ke2017lightgbm} and XGBoost \cite{chen2016xgboost} models.
    
    \textbf{Support Vector Machines}: Support vector machines \cite{hearst1998support} are fundamentally used for various tasks such as classification, regression or outlier detection. It builds a hyperplane or set of hyperplanes in a space of high or infinite dimensions. We use the kernel based on Radial Basis Function (RBF) \cite{park1991universal} for both the classification and regression tasks.
    
    \textbf{Bayesian Models}: Bayesian models are statistical models that use probabilities to represent not only the uncertainty about the output in the model but also the uncertainty about the input parameters to the model. We use a multivariate Bernoulli model \cite{dai2013multivariate} for the classification task and a ridge estimator \cite{li2010new} based Bayesian model for the regression task.

    \textbf{Deep Learning based Models}: For the Deep learning-based models, we use a Resnet 1D variant with 18 layers from \cite{patwa2021can}. The classifier models are trained using Binary cross-entropy as the loss function, while the regressor was trained using RMSE loss. Each of the models was trained for 500 epochs, and the best intermediate model was used for predicting the final popularity sequence.

We use the default hyper-parameter settings for all the ML-based models in our experiments. Table \ref{table:mainresultsCD} shows the list of all considered models and their performance in terms of truncated RMSE (mean, median) and truncated MAE (mean, median) for cases C \& D.

We observe that the tree-based models such as random forest, XGBoost are able to perform relatively better due to their robustness in dealing with data containing outliers. Random forest, in particular, has shown the best performance due to the usage of multiple trees for predictions. The Random forest model combined with the proposed settings, resulted in the best overall performance of tRMSE mean and median of 8.766 and 6.103 in case C, 9.05 and 6.38 in case D, showing significant improvements.


\section{Conclusion}
We studied the implications of time on the prediction of image popularity by applying dynamic weighting of feature sequences. Further, we proposed a novel strategy to mitigate the skewness in the distribution of a variable related to the sequence of features. We have also proposed a dynamic feature weighting technique to learn the importance of the feature sequences over time. We have conducted comprehensive experiments with the proposed framework on the standard Flickr dataset and compared the results with the state-of-the-art. The proposed dynamic feature weighting technique outperformed all the existing models. In future, semantic image features can be coupled with the proposed framework, for popularity prediction dynamics.


\bibliographystyle{ACM-Reference-Format}
\bibliography{references}

\section{Appendix}
\begin{figure*}
\begin{center}
  \includegraphics[width=\linewidth]{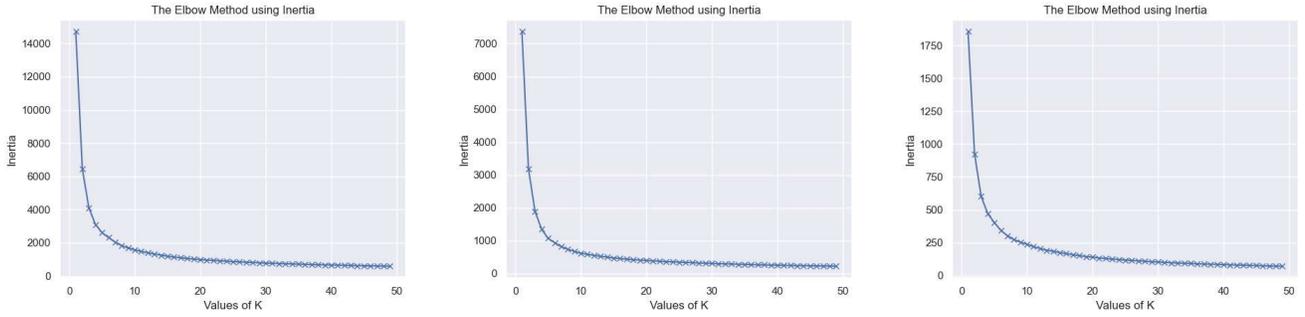}
\end{center}
\caption{K-means clustering elbow method graphs for K's range [1, 50]}
\label{fig:elbow_50}
\end{figure*}

\begin{figure*}
\begin{center}
  \includegraphics[width=\linewidth]{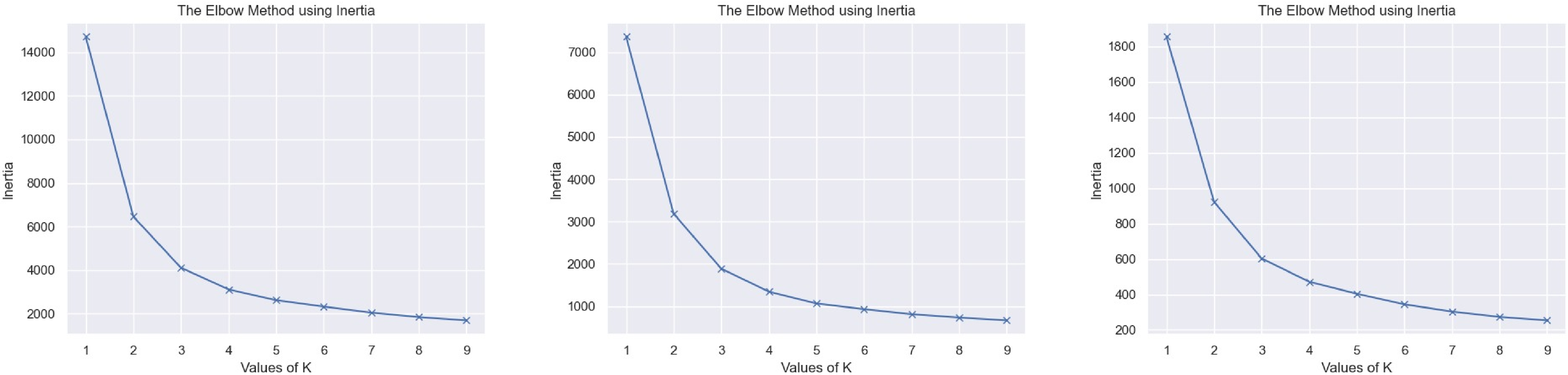}
\end{center}
\caption{K-means clustering elbow method graphs for K's range [1, 10]}
\label{fig:elbow_10}
\end{figure*}

As mentioned before, in order to discover the optimal value for $K$ in the K-means algorithm, we make use of the elbow method. Figure \ref{fig:elbow_50} shows the resulting graphs after plotting different values of K in the range of [1,50] against the Inertia while using the elbow method. Accordingly, Figure \ref{fig:elbow_10} shows the resulting graphs after plotting different values of K in the range of [1,10] against the Inertia while using the elbow method. 

We also include the tables for evaluation cases A \& B for the different experiments that were carried out in section \ref{ablation}.

\begin{table*}
\centering
\caption{The performance of the proposed framework in terms of the Case A \& B as compared to the existing normalization techniques ($s_{scale}$ normalization). Random Forest is used for all the evaluation settings.}
\label{table:normalizationresultsAB}
\begin{tabular}{|c|c|c|c|c|c|c|c|c|} 
\hline
\multirow{2}{*}{\begin{tabular}[c]{@{}c@{}}\textbf{Scaling}\\\textbf{Method}\end{tabular}} & \multicolumn{4}{c|}{\textbf{Case A}} & \multicolumn{4}{c|}{\textbf{Case B}} \\ 
\cline{2-9}
 & $tRMSE_{mean}$ & $tRMSE_{median}$ & $tMAE_{mean}$ & $tMAE_{median}$ & $tRMSE_{mean}$ & $tRMSE_{median}$ & $tMAE_{mean}$ & $tMAE_{median}$ \\ 
\hline
No scaling & 1.608±0.014 & 1.221±0.024 & 1.401±0.014 & 1.053±0.02 & 2.152±0.03 &   1.626±0.023  & 1.941±0.025 & 1.447±0.019 \\ 
\hline
log & 1.608±0.014 & 1.221±0.024 & 1.401±0.014 & 1.053±0.02 & 2.138±0.034 & 1.619±0.036 & 1.924±0.035 & 1.432±0.036 \\ 
\hline
\begin{tabular}[c]{@{}c@{}}log(scale/30 \\+ 0.1)\end{tabular} & 1.608±0.014 & 1.221±0.024 & 1.401±0.014 & 1.053±0.02 & 2.145±0.039 & 1.62±0.035 & 1.934±0.035 & 1.436±0.031 \\ 
\hline
Ortis et al. & 1.608±0.014 & 1.221±0.024 & 1.401±0.014 & 1.053±0.02 & 2.145±0.034 & 1.621±0.036 & 1.928±0.034 & 1.434±0.037 \\ 
\hline
\begin{tabular}[c]{@{}c@{}}Proposed\\ normalization \end{tabular}& 1.608±0.014\textbf{} & 1.221±0.024 & 1.401±0.014 & 1.053±0.02 & 2.146±0.023 & 1.624±0.029 & 1.934±0.018 & 1.442±0.02 \\
\hline
\end{tabular}
\end{table*}
\begin{table*}
\centering
\caption{The performance in terms of the Case A \& B metrics for different classification and regression modeling techniques}
\label{table:mainresultsAB}
\begin{tabular}{|c|c|c|c|c|c|c|c|c|} 
\hline
\multirow{2}{*}{\textbf{Model}} & \multicolumn{4}{c|}{\textbf{Case A}}                                & \multicolumn{4}{c|}{\textbf{Case B}}                                 \\ 
\cline{2-9}
                                & $tRMSE_{mean}$ & $tRMSE_{median}$ & $tMAE_{mean}$ & $tMAE_{median}$ & $tRMSE_{mean}$ & $tRMSE_{median}$ & $tMAE_{mean}$ & $tMAE_{median}$  \\ 
\hline
XGBoost                         & 1.608±0.014~   & ~1.221±0.024     & 1.401±0.014   & 1.053±0.02~     & ~2.217±0.027   & ~1.66±0.029      & ~2.003±0.024  & 1.479±0.023~     \\ 
\hline
LightGBM                        & 1.608±0.014~   & ~1.221±0.024     & 1.401±0.014   & 1.053±0.02~     & 2.196±0.038    & 1.64±0.032       & 1.983±0.035   & 1.462±0.03       \\ 
\hline
Bayesian                       & 1.608±0.014    & ~1.221±0.024~    & ~1.401±0.014  & 1.053±0.02      & 2.304±0.039~   & 1.722±0.037~     & 2.087±0.028   & ~1.535±0.028~    \\ 
\hline
SVM                            & ~1.608±0.014   & ~1.221±0.024~    & 1.401±0.014   & 1.053±0.02~     & ~2.272±0.041   & 1.684±0.032      & 2.059±0.033   & 1.504±0.023~     \\ 
\hline
Resnet 1D                         & 1.608 ± 0.025         & 1.221 ± 0.026            & 1.401 ± 0.023        & 1.053±0.02~            & 2.295 ± 0.047         & 1.703 ± 0.038            & 2.078 ± 0.047        & 1.518 ± 0.034             \\ 
\hline
RF                             & 1.608±0.014~   & ~1.221±0.024~~   & 1.401±0.014   & 1.053±0.02      & 2.146±0.023~   & ~1.624±0.029     & 1.934±0.018   & 1.442±0.02~      \\
\hline
\end{tabular}
\end{table*}

\begin{table*}[!t]
\centering
\caption{The performance in terms of the Case A \& B metrics for different time period lengths}
\label{table:caseABtime}
\begin{tabular}{|c|c|c|c|c|c|c|c|c|}
\hline
\multirow{2}{*}{\textbf{Period Length}} & \multicolumn{4}{c|}{\textbf{Case A}} & \multicolumn{4}{c|}{\textbf{Case B}} \\ \cline{2-9} 
 & $tRMSE_{mean}$ & $tRMSE_{median}$ & $tMAE_{mean}$ & $tMAE_{median}$ & $tRMSE_{mean}$ & $tRMSE_{median}$ & $tMAE_{mean}$ & $tMAE_{median}$ \\ \hline
5 & 1.451±0.023 & 1.089±0.019 & 1.343±0.023 & 0.998±0.015 & 1.978±0.046 & 1.482±0.033 & 1.876±0.048 & 1.39±0.032 \\ \hline
10 & 1.608±0.014 & 1.221±0.024 & 1.401±0.014 & 1.053±0.02 & 2.153±0.037 & 1.625±0.033 & 1.941±0.033 & 1.441±0.028 \\ \hline
15 & 1.713±0.011 & 1.293±0.09 & 1.439±0.01 & 1.079±0.007 & 2.284±0.005 & 1.731±0.003 & 1.983±0.005 & 1.491±0.0 \\ \hline
30 & 1.948±0.049 & 1.499±0.032 & 1.509±0.033 & 1.138±0.021 & 2.689±0.073 & 2.044±0.031 & 2.083±0.05 & 1.573±0.028 \\ \hline
\end{tabular}
\end{table*}

\begin{table*}
\centering
\caption{The performance of the proposed normalization technique($s_{scale}$ normalization) in terms of the Case A \& B metrics for different values of c. Random Forest is used for all the evaluation settings.}
\label{table:normalizationresultsforc2AB}
\begin{tabular}{|c|c|c|c|c|c|c|c|c|} 
\hline
\multirow{2}{*}{\begin{tabular}[c]{@{}c@{}}\textbf{Scaling}\\\textbf{Method}\end{tabular}} & \multicolumn{4}{c|}{\textbf{Case A}}                                                                  & \multicolumn{4}{c|}{\textbf{Case B}}                                                                                                                                           \\ 
\cline{2-9}
                                                                                            & $tRMSE_{mean}$         & $tRMSE_{median}$     & $tMAE_{mean}$         & $tMAE_{median}$     & $tRMSE_{mean}$         & $tRMSE_{median}$     & $tMAE_{mean}$        & $tMAE_{median}$                                                                           \\ 
\hline

c = 0.5 &1.608±0.014   &  1.221±0.024  & 1.401±0.014  &  1.053±0.02 &  2.144±0.03  &  1.623±0.034  &1.933±0.024 & 1.44±0.025                                                       \\ 
\hline

c = 0.1                                                                                    & 1.608±0.014  & 1.221±0.024  & 1.401±0.014  & 1.053±0.02 &

 2.147±0.036  & 1.625±0.034 & 1.935±0.031 & 1.445±0.029                                                                                   \\ 
\hline
c = one & 1.608±0.014 & 1.221±0.024 & 1.401±0.014 & 1.053±0.02 & 2.145±0.025 & 1.622±0.031 & 1.934±0.021 & 1.442±0.027 \\
\hline 

\begin{tabular}[c]{@{}c@{}}log(scale/30 \\+ 0.1)\end{tabular} &  1.608±0.014 & 1.221±0.024 & 1.401±0.014 & 1.053±0.02 & 2.145±0.039 & 1.62±0.035 & 1.934±0.035 & 1.436±0.031 \\ 
\hline
\begin{tabular}[c]{@{}c@{}}log(scale/30\\+ 0.5)\end{tabular} & 1.608±0.014 & 1.221±0.024 & 1.401±0.014 & 1.053±0.02 & 2.155±0.029 & 1.622±0.032 & 1.943±0.025 & 1.441±0.025 \\ 
\hline
 \begin{tabular}[c]{@{}c@{}}log(scale/30\\+ 1) \end{tabular} & 1.608±0.014 & 1.221±0.024 & 1.401±0.014 & 1.053±0.02 & 2.145±0.034 & 1.621±0.036 & 1.928±0.034 & 1.434±0.037 \\ 
\hline

\end{tabular}
\end{table*}

\begin{table*}
\centering
\caption{The performance in terms of the Case A \& B metrics for different combinations of cluster sizes}
\label{table:mainresultsAB_}
\begin{tabular}{|c|c|c|c|c|c|c|c|c|} 
\hline
\multirow{2}{*}{ \textbf{Cluster Size} } & \multicolumn{4}{c|}{\textbf{Case A} } & \multicolumn{4}{c|}{\textbf{Case B} } \\ 
\cline{2-9}
 & $tRMSE_{mean}$  & $tRMSE_{median}$  & $tMAE_{mean}$  & $tMAE_{median}$  & $tRMSE_{mean}$  & $tRMSE_{median}$  & $tMAE_{mean}$  & $tMAE_{median}$  \\ 
\hline
(2, 2, 2) & 1.608±0.014 & 1.221±0.024 & 1.401±0.014 &  1.053±0.02 & 2.146±0.023   &   1.624±0.029 & 1.934±0.018 &  1.442±0.02 \\
\hline
(2, 2, 3) &	1.56±0.013 & 1.185±0.022 & 1.36±0.014 & 1.02±0.018 &	2.134±0.038 & 1.61±0.029 & 1.922±0.033 & 1.423±0.024 \\
\hline
(3, 3, 3) &	1.31±0.013 & 0.994±0.015 & 1.12±0.013 & 0.835±0.011	& 2.066±0.008 & 1.573±0.007 & 1.85±0.008 & 1.386±0.002 \\
\hline
(3, 3, 4) &	1.281±0.017 & 0.973±0.018 & 1.096±0.016 & 0.818±0.014 &	2.066±0.017 & 1.572±0.007 & 1.846±0.017 & 1.384±0.004 \\
\hline

(3, 2, 4) & 1.39±0.017 & 1.06±0.019 & 1.198±0.017 & 0.9±0.014 & 2.047±0.011 & 1.562±0.008 & 1.828±0.009 & 1.37±0.003 \\

\hline
(3, 2, 3) &	1.419±0.013 & 1.08±0.016 & 1.222±0.013 & 0.917±0.01 & 2.045±0.012 & 1.555±0.01 & 1.829±0.01 & 1.37±0.004 \\

\hline
(3, 3, 5) &  1.27±0.013 & 0.963±0.012 & 1.086±0.012 & 0.809±0.01 & 2.062±0.01 & 1.57±0.015 & 1.841±0.006 & 1.377±0.009 \\																			
\hline
\end{tabular}
\end{table*}



\end{document}